\documentclass[conference]{IEEEtran}
\IEEEoverridecommandlockouts
\usepackage{cite}
\usepackage{amsmath,amssymb,amsfonts}
\usepackage{algorithmic}
\usepackage{graphicx}
\usepackage{textcomp}
\usepackage{xcolor}
\usepackage{multirow}
\usepackage{array}
\def\BibTeX{{\rm B\kern-.05em{\sc i\kern-.025em b}\kern-.08em
    T\kern-.1667em\lower.7ex\hbox{E}\kern-.125emX}}
\begin{document}

\title{Constrained PSO Six-Parameter Fuzzy PID Tuning Method for Balanced Optimization of Depth Tracking Performance in Underwater Vehicles
\thanks{Identify applicable funding agency here. If none, delete this.}
}

\author{\IEEEauthorblockN{1\textsuperscript{st} Yanxi Ding}
\IEEEauthorblockA{\textit{School of Engineering} \\
\textit{China University of Petroleum-Beijing at Karamay}\\
Karamay, China \\
2023015999@st.cupk.edu.cn}
\and
\IEEEauthorblockN{2\textsuperscript{nd} Tingyue Jia}
\IEEEauthorblockA{\textit{School of Engineering} \\
\textit{China University of Petroleum-Beijing at Karamay}\\
Karamay, China \\
2023016161@st.cupk.edu.cn}
}

\maketitle

\begin{abstract}
Depth control of underwater vehicles in engineering applications must simultaneously satisfy requirements for rapid tracking, low overshoot, and actuator constraints. Traditional fuzzy PID tuning often relies on empirical methods, making it difficult to achieve a stable and reproducible equilibrium solution between performance enhancement and control cost. This paper proposes a constrained particle swarm optimization (PSO) method for tuning six-parameter fuzzy PID controllers. By adjusting the benchmark PID parameters alongside the fuzzy controller's input quantization factor and output proportional gain, it achieves synergistic optimization of the overall tuning strength and dynamic response characteristics of the fuzzy PID system. To ensure engineering feasibility of the optimization results, a time-weighted absolute error integral, adjustment time, relative overshoot control energy, and saturation occupancy rate are introduced. Control energy constraints are applied to construct a constraint-driven comprehensive evaluation system, suppressing pseudo-improvements achieved solely by increasing control inputs. Simulation results demonstrate that, while maintaining consistent control energy and saturation levels, the proposed method significantly enhances deep tracking performance: the time-weighted absolute error integral decreases from 0.2631 to 0.1473, the settling time shortens from 2.301 s to 1.613 s, and the relative overshoot reduces from 0.1494 to 0.01839. Control energy varied from 7980 to 7935, satisfying the energy constraint, while saturation occupancy decreased from 0.004 to 0.003. These results validate the effectiveness and engineering significance of the proposed constrained six-parameter joint tuning strategy for depth control in underwater vehicle navigation scenarios.
\end{abstract}


\section{Introduction}

Underwater vehicles typically require stable, rapid, and safe depth maintenance and descent capabilities for tasks such as ocean observation, resource exploration, and engineering operations. The challenges in depth control stem from: complex hydrodynamic environments, significant system inertia and lag, and the difficulty in accurately modeling external disturbances \cite{2}. Concurrently, actuators face amplitude constraints and energy consumption limitations. Against this backdrop, controller tuning must not only enhance tracking speed and steady-state accuracy but also suppress overshoot and oscillations. This prevents saturation and engineering unfeasibility risks arising from excessive control inputs.

Fuzzy PID can utilize information such as error and its rate of change to perform online parameter tuning, thereby mitigating the reliance of traditional PID on precise models and manual tuning experience to a certain extent. Consequently, it is widely applied in the control of complex nonlinear and uncertain systems. However, for deep control tasks constrained by engineering limitations, fuzzy PID tuning still faces two prominent challenges: First, controller parameters exhibit high sensitivity to closed-loop performance, making it difficult to achieve a balanced performance that simultaneously prioritizes speed and smoothness through empirical tuning; Second, if error-based metrics are the sole optimization objective, intelligent tuning may sacrifice apparent performance by increasing control inputs, leading to elevated control costs or actuator operation in nonlinear saturation regions, thereby compromising deployability and repeatability \cite{3}. Consequently, achieving systematic tuning and performance-balanced optimization of fuzzy PID under constraints holds clear engineering value.

To address the aforementioned issues, this paper proposes a constrained particle swarm optimization (PSO)-based six-parameter fuzzy PID tuning method for depth tracking of underwater vehicles. The controller parameters are unified as six-dimensional optimization variables: the baseline PID parameters ${K}_{{p}_0}$, ${K}_{{i}_0}$, ${K}_{{d}_0}$, the input quantization factor ${K}_e$, ${K}_{{e}_c}$, and the output proportional factor ${K}_u$ of the fuzzy controller. PSO is employed to achieve coordinated tuning of these six parameters. Furthermore, considering engineering feasibility, this paper integrates transient quality and control cost into a unified evaluation framework. An energy constraint is introduced to suppress pseudo-improvements achieved through excessive control effort, ensuring performance gains while satisfying actuator limitations. Through comparative experiments on the Simulink simulation platform, the proposed method demonstrates improved depth tracking performance without significantly increasing control cost.

\section{Related Work}

\subsection{Depth Control and Engineering Constraints for Underwater Vehicles}

Underwater vehicles encounter significant nonlinearities, hydrodynamic parameter uncertainties, and external disturbance effects during depth control \cite{2}. For depth-holding and variable-depth missions, researchers typically extract the vertical plane dynamics channel based on a six-degree-of-freedom motion model under small-angle and steady-speed assumptions, designing corresponding feedback controllers. Due to common constraints such as amplitude saturation, rate limitations, and energy consumption on actuators like thrusters and rudders, depth control in engineering practice must not only ensure tracking accuracy and transient performance but also prevent excessive control inputs from causing saturation, oscillations, or significant energy consumption increases. Against this backdrop, achieving a reasonable trade-off between dynamic performance and actuator feasibility has become an unavoidable engineering challenge in underwater vehicle control research. However, some existing studies still primarily focus on error metrics during controller design or parameter tuning, with relatively limited attention to control costs and actuator constraints.

\subsection{Application of PID and Fuzzy PID in Underwater Vehicles}

PID control, due to its simple structure and proven engineering maturity, is widely applied in depth control for underwater vehicles. However, under conditions of parameter uncertainty and operational variations, fixed-parameter PID often struggles to simultaneously achieve both rapid response and stability across different operational phases \cite{4}. To enhance the controller's adaptability to nonlinearity and uncertainty, fuzzy PID control introduces error and its rate of change to perform online parameter adjustment, thereby reducing reliance on precise models and manual tuning. Existing research indicates that fuzzy PID offers advantages in suppressing overshoot and improving transient response. However, the overall performance of fuzzy PID is highly sensitive to the reference PID parameters, as well as the fuzzy quantization factor and output scaling factor. In most engineering applications, these parameters are still primarily selected based on experience, lacking systematic and reproducible tuning methods. This limitation restricts the stable performance of fuzzy PID in complex engineering systems.

\subsection{Research on Intelligent Optimization Methods for Fuzzy PID Parameter Tuning}

To overcome the limitations of empirical tuning, intelligent optimization methods such as genetic algorithms and particle swarm optimization are widely employed for PID and fuzzy PID parameter tuning. PSO offers advantages including fewer parameters, simple implementation, and strong continuous domain search capabilities, making it suitable for offline optimization of control parameters \cite{7}. Some studies have incorporated PSO into fuzzy PID design, optimizing some or all control parameters and verifying its effectiveness in improving dynamic performance \cite{10}.

\subsection{Limitations of Existing Research and Positioning of This Paper}
However, existing relevant research still exhibits two shortcomings: On one hand, some studies only optimize the benchmark PID parameters without uniformly modeling and jointly optimizing the quantization factor and output scaling factor within the fuzzy controller, making it difficult to fully demonstrate the overall regulation capability of fuzzy PID. Second, optimization objectives predominantly focus on error or response time, lacking explicit consideration of engineering constraints such as control energy and saturation behavior \cite{3}. This tendency leads to optimization results that rely on larger control inputs to achieve apparent performance improvements, thereby compromising engineering feasibility \cite{11}. To address these issues, this paper proposes a constrained PSO-based six-parameter fuzzy PID tuning method for underwater vehicle depth tracking tasks, emphasizing engineering applicability and deployability. While preserving the fuzzy control structure and rule base, it unifies the benchmark PID parameters and key scaling factors of the fuzzy controller into six-dimensional optimization variables, achieving holistic collaborative tuning via particle swarm optimization. Furthermore, this paper moves beyond solely pursuing minimal error or fastest response. Instead, it constructs a constrained evaluation system that balances tracking performance, transient quality, and actuator load \cite{12}. By introducing constraints on control energy and saturation occupancy rate, the optimization process is guided to achieve a reasonable equilibrium between performance enhancement and control cost. This approach aims to avoid pseudo-improvements where greater control effort yields only apparent performance gains, ensuring the resulting parameters better align with the engineering application requirements of underwater vehicles.

\section{Establishment of an Underwater Vehicle Model}

Underwater vehicle is typically categorized by mission and control method into ROV (Remotely Operated Vehicle) and AUV (Autonomous Underwater Vehicle). The primary difference between them lies in mission organization and the degree of human involvement at the control level. However, their modeling frameworks for six-degree-of-freedom rigid-body motion under hydrodynamic forces are unified. Under specific operating conditions, depth channels can be extracted for controller comparison studies. To characterize the vertical-plane depth-varying dynamics of underwater vehicles, establish a fixed coordinate system and a body coordinate system, with the origin of the body coordinate system located at the vehicle's center of gravity. Considering small-angle conditions during typical depth-changing processes, under the assumptions of varying depth only without altering heading, neglecting roll, and assuming negligible coupling between the vertical and horizontal planes, steady-state linear motion in the vertical plane can be linearized near equilibrium pitch angle and rudder angle \cite{1}. This yields the linearized system of equations for vertical velocity and longitudinal pitch rate:
\begin{equation} \label{eq:1}
	\left\{\begin{aligned}
		&\left(m-Z_{\omega^{\prime}}\right)\dot{\omega}-Z_{q^{\prime}}\dot{q}-\left(mV+Z_q\right)q-Z_{\omega}\omega=T_z \\
		&\left(I_y-M_{q^{\prime}}\right)\dot{q}-M_{\omega^{\prime}}\dot{\omega}-M_qq-M_\omega\omega=M_{T_z}
	\end{aligned}\right.
\end{equation}

Furthermore, based on the small-angle approximation and neglecting higher-order hydrodynamic terms, Eq.\eqref{eq:2} can be rearranged and the intermediate variable eliminated in the Laplace domain, yielding a fourth-order transfer function from rudder angle to depth:
\begin{equation} \label{eq:2}
	G\left(s\right)=\frac{z\left(s\right)}{\delta_e\left(s\right)}=\frac{Z_{\delta_e}V\left(I_ys^2-M_qs+mgh\right)}{A_3s^4+A_2s^3+A_1s^2+A_0s}
\end{equation}

The coefficients of the denominator polynomial are:
\begin{equation} \label{eq:3}
	\left\{\begin{aligned}
		A_0&=-mgh \\
		A_1&=M_qZ_\omega+mgh\left(m-Z_\omega\right)-M_\omega\left(mV+Z_q\right) \\
		A_2&=-M_q\left(m-Z_\omega\right)-I_yZ_{\omega^{\prime}} \\
		A_3&=I_y\left(m-Z_{\omega^{\prime}}\right)
	\end{aligned}\right.
\end{equation}

This paper selects a specific type of underwater vehicle carrier as the simulation research subject. By substituting the carrier model parameters and various hydrodynamic parameters into the transfer function formula, the transfer function for the AUV's depth motion is obtained as:
\begin{equation} \label{eq:4}
	G\left(s\right)=\frac{z}{\delta_e}=\frac{0.3559s^2+5.226s+35.2459}{s^4+10.0997s^3+8.3879s^2}
\end{equation}

The information of relevant symbols in the equation can be found in Table \ref{tab:1}.

\begin{table}[!ht]
	\centering
	\caption{Description of Symbols\label{tab:1}}
	\begin{tabular}{cm{5.5cm}<{\centering}}
		\hline
		\textbf{Symbol}	&	\textbf{Physical meaning} \\
		\hline
		$z$	&	Depth (vertical displacement) \\
		$\delta_e$	&	Horizontal Rudder Angle \\
		$\omega$	&	Vertical velocity (along the z-axis), $\omega=\dot{z}$ \\
		$q$	&	Longitudinal angular velocity, $q=\dot{\theta}$ \\
		$\theta$	&	Longitudinal inclination (small angle) \\
		$m$	&	Vehicle mass \\
		$I_y$	&	Moment of inertia about y-axis \\
		$V$	&	Travel speed \\
		$g$	&	Gravitational acceleration \\
		$h$	&	Distance between center of gravity and center of buoyancy (restoring force arm) \\
		$Z_\omega,Z_{\omega^{\prime}},Z_q,Z_{q^{\prime}}$	&	Vertical force hydrodynamic derivative \\
		$M_\omega,M_{\omega^{\prime}},M_q,M_{q^{\prime}}$	&	Pitching Moment Hydrodynamic Derivative \\
		$Z_{\delta_e}$	&	Derivative of rudder effect with respect to vertical force at rudder angle \\
		\hline
	\end{tabular}
\end{table}

\section{Design of the Controller and Fitness Function}

\begin{figure}[!ht]
	\centering
	\includegraphics[width=\linewidth]{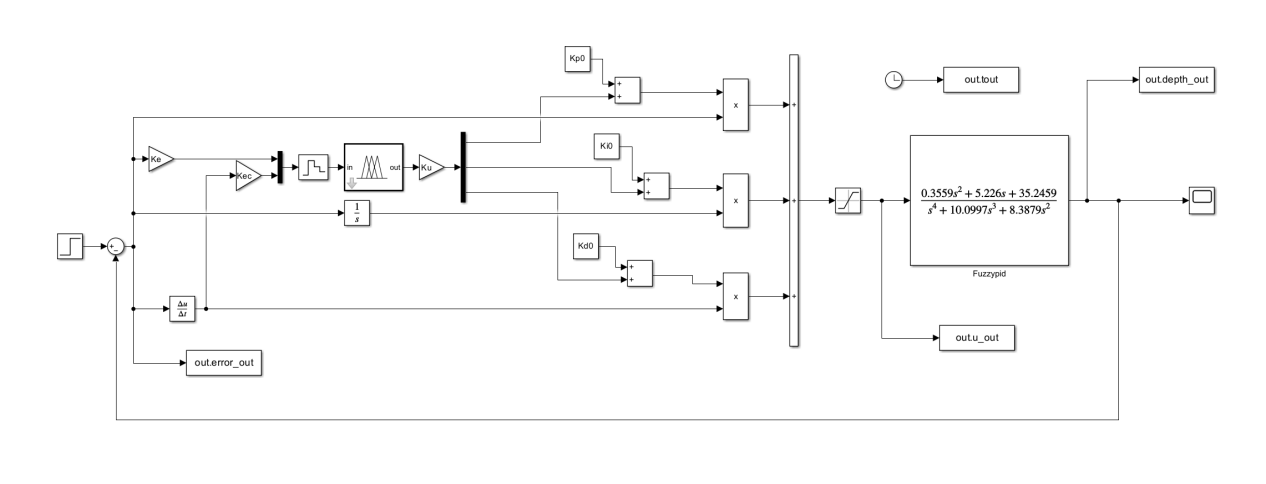}
	\caption{Controller structure.}
	\label{fig:1}
\end{figure}

The fuzzy PID controller for underwater vehicles described in this paper adopts a two-input, three-output structure. The inputs to the fuzzy controller are the error and the corresponding error rate of change, while the outputs represent the correction values $\delta{K}_p$, $\delta{K}_i$, $\delta{K}_d$ for the PID controller parameters. During system operation, input values($e$ and ${e}_c$) undergo fuzzification, fuzzy reasoning, and defuzzification. The resulting online correction values($\delta{K}_p$, $\delta{K}_i$, $\delta{K}_d$) are fed into the PID controller, enabling the underwater vehicle to dynamically adjust parameters(${K}_p$, ${K}_i$, ${K}_d$) in real time across various motion states to ensure the system achieves optimal responsiveness and the underwater vehicle maintains excellent static and dynamic performance. The structure is shown in the Fig. \ref{fig:1}.

\subsection{Fuzzy Input Quantity}

\begin{figure}[!ht]
	\centering
	\includegraphics[width=0.8\linewidth]{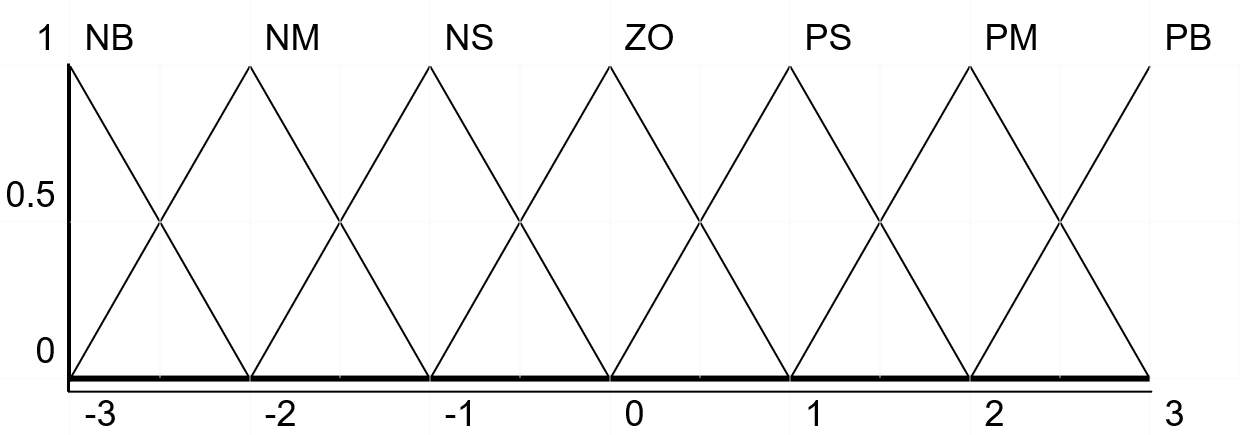}
	\caption{Membership function.}
	\label{fig:2}
\end{figure}

The purpose of the fuzzification process is to map precise input values onto a predefined fuzzy domain for membership calculation and rule inference \cite{5}. Fuzzification determines the membership degree of precise input values based on the membership function of fuzzy sets. Input quantities undergo normalization processing, and the input domain undergoes discretization processing, transforming into corresponding variable values within the domain. Fuzzy rules adopt the standardized design created by Mamdani, defining the ranges of inputs and outputs as domains over fuzzy sets [-3,3]. Select seven fuzzy subsets, where $N$ denotes negative, $P$ denotes positive, $B$ denotes large, $M$ denotes medium, $S$ denotes small, and $ZO$ denotes zero \cite{6}. The membership functions for all fuzzy subsets adopt the commonly used triangular form, with the membership function for error shown in the Fig. \ref{fig:2}.

\subsection{Establishment of Fuzzy Control Rules}

Establishing fuzzy control rules is the most critical step in designing a fuzzy PID controller. Considering the hydrodynamic characteristics and control requirements of the underwater vehicle, appropriate fuzzy statements can be selected to establish a three-output($\delta{K}_p$, $\delta{K}_i$, $\delta{K}_d$) fuzzy control rule table, as shown in the Table \ref{tab:2}. The fuzzy control rule format is: If input Then output \cite{6}.

\begin{table*}[!ht]
	\centering
	\caption{Model comparison\label{tab:2}}
	\begin{tabular}{ccccccccc}
		\hline
		\multicolumn{2}{c}{\multirow{2}{*}{$\boldsymbol{{\delta}{K}_{p}/{\delta}{K}_{i}/{\delta}{K}_{d}}$}} & \multicolumn{7}{c}{$\boldsymbol{e_c}$} \\
		\cline{3-9}
		&	&	\textbf{\textit{NB}}	&	\textbf{\textit{NM}}	&	\textbf{\textit{NS}}	&	\textbf{\textit{ZO}}	&	\textbf{\textit{PS}}	&	\textbf{\textit{PM}}	&	\textbf{\textit{PB}} \\
		\hline
		\multirow{7}{*}{$\boldsymbol{e}$}	&	\textbf{\textit{NB}}	&	PB/NB/PS	&	PB/NB/NS	&	PM/NM/NB	&	PM/NM/NB	&	PS/NS/NB	&	ZO/ZO/NM	&	ZO/ZO/PS	\\
		&	\textbf{\textit{NM}}	&	PB/NB/ZO	&	PB/NM/NS	&	PM/NS/NM	&	PS/NS/NM	&	PS/ZO/NS	&	ZO/PS/NS	&	NS/PS/ZO	\\
		&	\textbf{\textit{NS}}	&	PM/NM/ZO	&	PM/NM/NS	&	PM/NS/NS	&	PS/ZO/NS	&	ZO/PS/NS	&	NS/PM/NS	&	NS/PM/ZO	\\
		&	\textbf{\textit{ZO}}	&	PM/NM/ZO	&	PM/NS/ZO	&	PS/ZO/ZO	&	ZO/PS/ZO	&	NS/PS/ZO	&	NM/PM/ZO	&	NM/PB/ZO	\\
		&	\textbf{\textit{PS}}	&	PS/ZO/PB	&	PS/ZO/NS	&	ZO/PS/PS	&	NS/PS/PS	&	NS/PM/PS	&	NM/PB/PS	&	NM/PB/PB	\\
		&	\textbf{\textit{PM}}	&	PS/ZO/PB	&	ZO/ZO/PM	&	NS/PS/PM	&	NM/PM/PM	&	NM/PM/PS	&	NM/PB/PS	&	NB/PB/PB	\\
		&	\textbf{\textit{PB}}	&	ZO/NB/PS	&	ZO/NB/NS	&	NM/NM/NB	&	NM/NM/NB	&	NM/NS/NB	&	NB/ZO/NM	&	NB/ZO/PS	\\
		\hline
	\end{tabular}
\end{table*}

\subsection{Output Quantity Defuzzification}

The defuzzification process converts fuzzy terms into quantized results required for executing the final control element, serving as a critical stage in Mamdani fuzzy inference systems. Defuzzification employs the centroid method \cite{15}. Let $A$ be a non-empty fuzzy set, discretized into m vertical slices at ${x}_1,\; {x}_2,\;\ldots,\; {x}_m$. By taking the weighted average of each element in the fuzzy control variable and its corresponding membership degree, the centroid of $A$ is given by Eq. \eqref{eq:5}. In Eq. \eqref{eq:5}, ${x}_A$ represents the precise value obtained through defuzzification, ${x}_i$ denotes the fuzzy variable element, and $\mu_A\left({x}_i\right)$ indicates the membership degree of element ${x}_i$.
\begin{equation} \label{eq:5}
	x_A = \frac{\sum_{i=1}^{m}\mu_A\left(x_i\right)x_i}{\sum_{i=1}^{m}{\mu_A\left(x_i\right)}}
\end{equation}

PID parameters are updated online using a reference value plus a correction term, as shown in Eq. \eqref{eq:6}.
\begin{equation} \label{eq:6}
	\left\{\begin{aligned}
		& K_p=K_{p_0}+\delta K_p \\
		& K_i=K_{i_0}+\delta K_i \\
		& K_d=K_{d_0}+\delta K_d
	\end{aligned}\right.
\end{equation}

\subsection{PSO Tuning Variables and Offline Optimization Framework}

Under the premise that the control structure and rule library remain unchanged, this paper employs particle swarm optimization for offline tuning of the controller's key parameters. The optimization variable is defined as $\Theta=\left\{{K}_{{P}_0},{K}_{{i}_0},{K}_{{d}_0},{K}_e,{K}_{{e}_c},{K}_u\right\}$. The first three parameters constitute the PID reference parameters, while the latter three represent the error, error rate of change, and output scaling factor (corresponding to the three gain modules in the Simulink block diagram). Each iteration of the PSO generates a set of candidate parameters, invokes a closed-loop simulation to obtain the time series of $e\left(t\right)$, $z\left(t\right)$ and $u\left(t\right)$, and evaluates this parameter set using the fitness function to update the particle swarm \cite{9}.

\subsection{Fitness Function and Evaluation Metrics}

To comprehensively evaluate control performance across three dimensions---tracking accuracy, dynamic quality, and actuator load---this paper employs five metrics: time-weighted absolute error integral (${J}_{ITAE}$), settling time (${T}_s$), overshoot ($OS$), control energy (${E}_u$), and saturation rate (${S}_r$, the proportion of time spent near the limit) \cite{14}. To mitigate the sensitivity of weighting to dimensional differences among metrics, ITAE and control energy are normalized, while settling time undergoes scaling. Additionally, to prevent the optimization process from excessively pursuing zero overshoot at the expense of response speed, a soft constraint is applied to overshoot using a tolerance interval with an excess penalty. Specifically, 2\% overshoot is permitted without penalty. This approach encourages the search to favor comprehensive compromises rather than single, overly conservative solutions. To ensure controllable actuator load, the optimized control energy is constrained not to increase significantly (2\% tolerance) relative to the baseline energy \cite{13}. Violations incur a strong penalty term (${E}_{pen}$), ensuring optimization prioritizes improving error dynamics over achieving apparent speed through larger control outputs \cite{8}. This mechanism embodies the engineering significance of constrained tuning. Based on the above, a weighted scalarized composite fitness function is employed:
\begin{equation} \label{eq:7}
	J\left(\Theta\right)=1.0J_{ITAE}+0.10E_u+1.2OS+0.6T_s+3.0S_r+E_{pen}
\end{equation}

\section{Experiments and Results}

\subsection{Experimental Setup and Comparison Plan}

To validate the effectiveness of the proposed PSO-optimized fuzzy PID control method for depth control in underwater vehicles, a comparative scheme was designed under identical simulation platforms and object models. Traditional PID employs manually set PID parameters (${K}_{{p}_0}=300$, ${K}_{{I}_0}=0.5$, ${K}_{{d}_0}=35$), while conventional fuzzy PID uses initial values identical to traditional PID. The proposed method maintains the fuzzy PID structure and rules unchanged, utilizing PSO for offline tuning of six critical parameters. All experiments employ identical commands, identical step depth commands, identical limiting constraints, and identical simulation durations.

\subsection{Comparative Analysis of Deep Tracking Responses}

\begin{figure}[!ht]
	\centering
	\includegraphics[width=\linewidth]{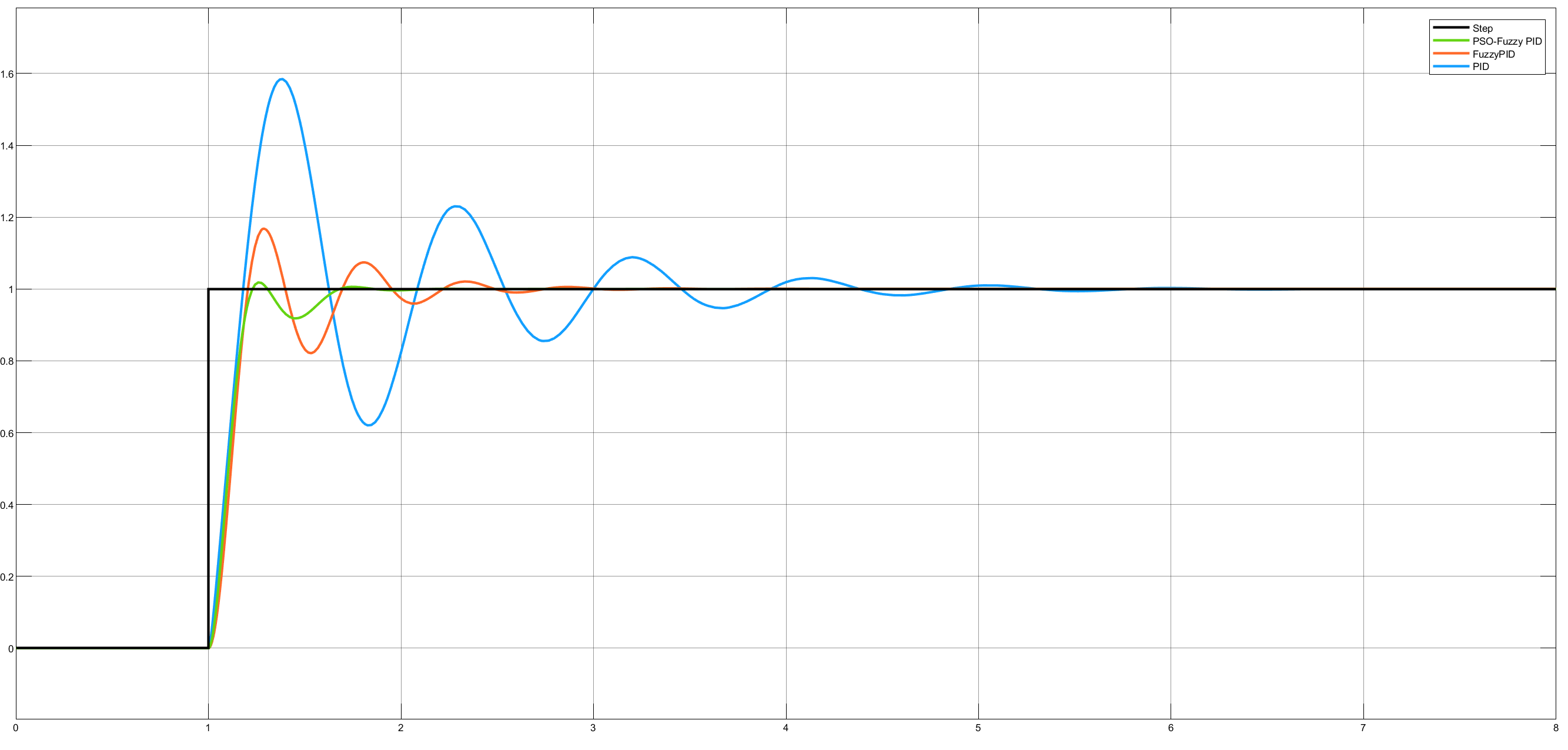}
	\caption{Response for deep tracking.}
	\label{fig:3}
\end{figure}

Fig. \ref{fig:3} presents the step response curves comparing three control strategies: traditional PID, fuzzy PID, and PSO-optimized fuzzy PID. The figure reveals that traditional PID exhibits significant overshoot during the initial response phase. The introduction of fuzzy PID enhances the system's adaptability across different error stages. The settling time of the PSO-optimized system decreased from 2.328 s to 1.612 s, a reduction of approximately 30.7\%, indicating faster dynamic response while maintaining stability. The fuzzy PID exhibits noticeable overshoot ($OS$ = 0.1548), whereas the PSO-optimized system reduces overshoot to 0.01413, representing a decrease exceeding 90\%. These results demonstrate that the optimized parameter configuration effectively suppresses overshoot during the initial response phase. Both control methods accurately track the target depth during the steady-state phase without introducing steady-state error, indicating that the optimization process did not compromise steady-state performance. This outcome demonstrates that PSO does not merely reduce overshoot by slowing the system but achieves a balanced trade-off between speed and stability through comprehensive parameter tuning. The PSO-optimized fuzzy PID configuration strikes a more optimal equilibrium among response speed, overshoot suppression, and control energy.

\subsection{Analysis of Control Input and Actuator Load}

\begin{figure}[!ht]
	\centering
	\includegraphics[width=\linewidth]{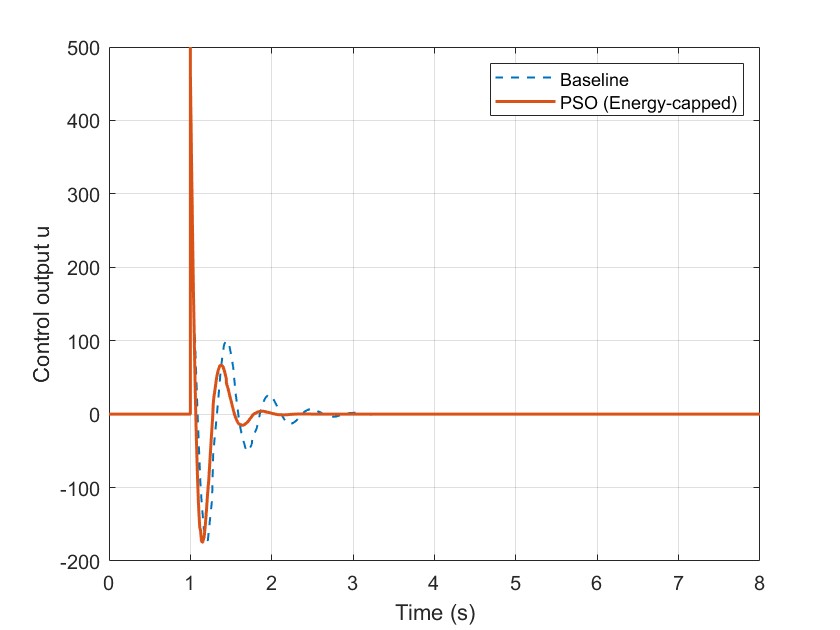}
	\caption{Control input.}
	\label{fig:4}
\end{figure}

The Fig. \ref{fig:4} shows the time-domain response of control input $u(t)$ for both the traditional fuzzy PID controller and the proposed constrained PSO six-parameter fuzzy PID controller under a step command, evaluating the impact of different control strategies on actuator drive intensity and saturation risk. It can be observed that during the initial phase of command abrupt changes, both control strategies generate substantial control inputs to overcome the system's initial error. Among them, the PSO-optimized scheme exhibits more concentrated control action during the transient phase, with its control input reaching a high amplitude within an extremely short time before rapidly decaying and approaching zero. In contrast, the baseline controller maintains multiple positive and negative oscillations over a longer period, with a slower decay process of the control input. Although the PSO-optimized scheme exhibits larger short-term control peaks during the initial transient phase, their duration is brief. Consequently, its overall control energy and saturation utilization do not significantly exceed those of the baseline scheme. This result indicates that the proposed method achieves a reasonable temporal distribution of control actions through optimized parameter configuration. It enhances dynamic response performance while effectively suppressing increased actuator load, demonstrating the practical significance of constrained tuning strategies in engineering applications.

\subsection{Analysis of PSO Convergence Characteristics}

\begin{figure}[!ht]
	\centering
	\includegraphics[width=\linewidth]{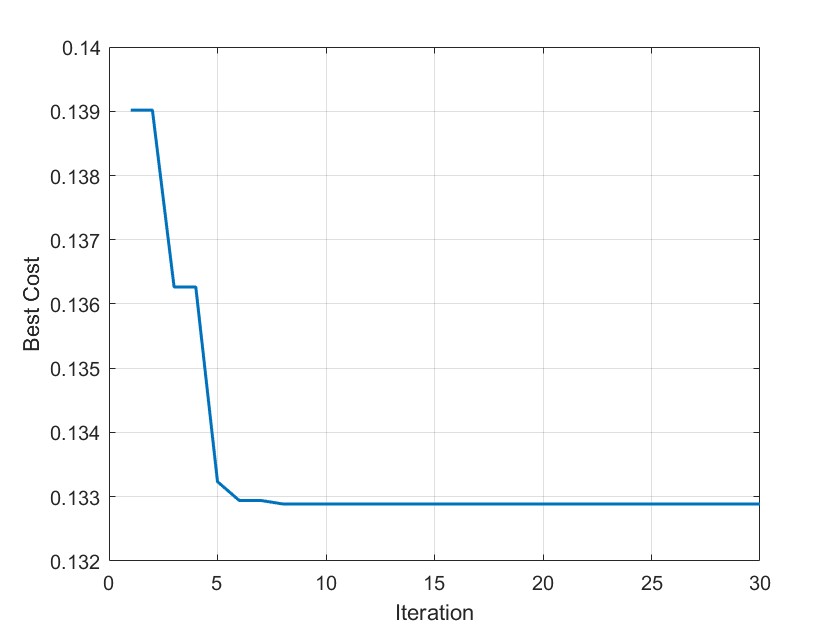}
	\caption{PSO convergence.}
	\label{fig:5}
\end{figure}

The Fig. \ref{fig:5} illustrates the evolution of the optimal fitness under fixed random seeds as iterations progress. A clear and stable convergence trend is evident: the optimal fitness declines rapidly during the initial iterations before gradually stabilizing without significant oscillations. This indicates that the fitness function is appropriately constructed and the search process is stable. It demonstrates that particle swarm optimization can effectively locate superior parameter combinations even under energy constraints.

\section{Conclusions}

\subsection{Conclusions of the Experiment}

This paper investigates a PSO-constrained optimization-based fuzzy PID parameter tuning method for depth control of underwater vehicles. While maintaining the existing fuzzy control structure and rules, the approach employs particle swarm optimization for offline tuning of critical parameters, explicitly constraining control energy and saturation behavior within the fitness function. Experimental results demonstrate that the PSO-optimized fuzzy PID achieves significant improvements in key dynamic performance metrics compared to the traditional fuzzy PID. Specifically, the settling time was reduced by 29.90\%, overshoot decreased by 87.69\%, while control energy and saturation occupancy remained at similar or even slightly lower levels. This indicates that the performance enhancement was not achieved by increasing control outputs. By incorporating control energy into the optimization objective, this study avoids the common pitfall in traditional intelligent optimization methods where apparent performance gains come at the expense of increased actuator burden. Experimental results validate that this constraint strategy effectively guides the search process, achieving a reasonable balance between dynamic response quality and actuator feasibility. It provides an effective approach for controller parameter tuning under constrained conditions.

\subsection{Analysis of Engineering Applications and Practical Scenarios}

From an engineering application perspective, the method proposed in this paper is particularly suitable for underwater operations that demand dynamic performance while being constrained by actuator capabilities and energy consumption. In seabed inspection, pipeline patrol, and underwater maintenance tasks, ROVs often require frequent small-amplitude depth adjustments. Excessive overshoot or control output can easily trigger attitude disturbances, thruster saturation, or even mechanical wear. This method significantly reduces overshoot without increasing control energy, demonstrating high engineering adaptability. For AUVs performing long-duration cruising or periodic depth-changing tasks, control energy directly impacts endurance and mission coverage. The proposed energy-constrained parameter tuning strategy suppresses unnecessary energy consumption while ensuring dynamic performance, offering clear practical value. In practical engineering, hydrodynamic parameters often vary with payload, speed, or environmental conditions. This method combines fuzzy structures with offline parameter optimization to enhance parameter configuration rationality without altering the control structure. It is suitable for industrial control systems with imperfectly accurate models or requiring rapid deployment.

\subsection{Future Work Outlook}

Although the proposed method has demonstrated its effectiveness in simulation environments, there remains scope for further research and extension. Subsequent studies could systematically evaluate the robustness and generalization capabilities of the proposed method under varying step magnitudes, initial depths, and multi-stage variable-depth commands. Additionally, introducing external disturbances such as waves, additional drag forces, or model parameter deviations could assess the applicability of the constrained optimization strategy in uncertain environments. When feasible, deploying the proposed method on hardware-in-the-loop platforms or actual underwater propulsion systems to validate its real-time performance and stability under real actuator and sensor conditions; Future work could extend the proposed constrained fitness function into a multi-objective optimization framework to further explore trade-offs between dynamic performance, energy consumption, and actuator lifespan.

\end{document}